\pdfoutput=1
\documentclass[11pt]{article}

\usepackage[]{acl}

\usepackage{times}
\usepackage{latexsym}

\usepackage{subcaption}
\usepackage{graphicx}

\usepackage[T1]{fontenc}
\usepackage[utf8]{inputenc}

\usepackage{microtype}

\title{Actionable Entities Recognition Benchmark for Interactive Fiction}

\author{
Alexey Tikhonov\\
Inworld.AI \\
Berlin, Germany \\
\texttt{altsoph@gmail.com}\And
Ivan P. Yamshchikov\\
Max Planck Institute for\\
 Mathematics in the Sciences\\
 Leipzig, Germany\\
 CEMAPRE, \\
 University of Lisbon, Portugal\\
  \texttt{ivan@yamshchikov.info} \\
}

\begin{document}
\maketitle
\begin{abstract}
 This paper presents a new natural language processing task - Actionable Entities Recognition (AER) - recognition of entities that protagonists could interact with for further plot development. Though similar to classical Named Entity Recognition (NER), it has profound differences. In particular, it is crucial for interactive fiction, where the agent needs to detect entities that might be useful in the future. We also discuss if AER  might be further helpful for the systems dealing with narrative processing since actionable entities profoundly impact the causal relationship in a story. We validate the proposed task on two previously available datasets and present a new benchmark dataset for the AER task that includes 5550 descriptions with one or more actionable entities. 
\end{abstract}

\noindent "One must never place a loaded rifle on the stage if it is not going to go off. It is wrong to make promises you do not mean to keep." 

A. Chekhov.

\section{Introduction}

One of the bottlenecks that hold modern Natural Language Processing (NLP) from the generation of longer texts is the concept of {\em narrative} or storyline. There are constant attempts to generate longer blocks of text, such as \cite{kedziorski2019understanding} or \cite{agafonova2020paranoid}. These attempts succeed under certain stylistic and topical constraints that exclude the problem of narrative generation altogether. Although there are several recent results in the areas of suspense generation \cite{doust2017model}, narrative personalization \cite{wang2017interactive}, and generation of short context-based narratives \cite{womack2019interactive},  generating long stories is still a challenge \cite{van2019narrative}.

Though philosophers and linguists have tried to conceptualize the notions of plot, narrative arc, action, and actor for almost a century \cite{shklovsky1925theory,propp1968morphology,van1976philosophy}, few of these theoretical concepts could be instrumental for modern NLP. \citet{ostermann2019mcscript2} present a machine comprehension corpus for the end-to-end evaluation of script knowledge. The authors demonstrate that though the task is not challenging to humans, existing machine comprehension models fail to perform well, even if they make use of a commonsense knowledge base. Despite these discouraging results, there are various attempts to advance narrative generation within the NLP community, see \cite{fan2019strategies,ammanabrolu2020story}.

 We believe that one of the possible ways to overcome the challenge of narrative generation could be found through more profound insights into a narrative structure. However, there are frustratingly few established NLP tasks that engage at least some aspects of natural language crucial for narrative generation and comprehension. Partly, that state of affairs could be attributed to the void in our conceptual understanding of narrative from the point of view of computer science, see \cite{yamshchikov2022wrong}. This paper tries to amend this and proposes a new NLP task — Actionable Entities Recognition (AER). The idea of the task is based on a simple premise: as the story unfolds reader understands which elements of the story might further affect the storyline. This phenomenon is especially pronounced in Interactive Fiction, where a player is trying to interact with the textually described environment via textual commands. We believe that detecting entities that the protagonists can interact with and that have an impact on the further development of the story might be instrumental for a better understanding of the story's structures and, ultimately, pave the way to entertaining generative fiction.  
 
The contributions of the paper are four-fold:
 
\begin{itemize}
    \item it formalizes the problem of Actionable Entities Recognition;
    \item it provides a dataset of AEs,  collected from various interactive fiction resources;
    \item it contrasts AER and NER tasks and demonstrates that modern language models can detect actionable entities in interactive fiction, and their predictions go in line with the decisions of human players;
    \item finally, it demonstrates that the frequency of AEs in the text corresponds with the macrostructure of a storyline. Thus AER task might be one of the stepping stones toward the generation of long entertaining stories.
\end{itemize}

\section{AER Task}

Actionable Entities Recognition (AER) is a subtask of information extraction that seeks to locate entities mentioned in an unstructured text that significantly affect the narrative as the story unfolds. Similarly to named entity recognition, AER works with the entities for which one or many strings, such as words or phrases, stand consistently for some referent in line with the so-called concept of {\em rigid designators}, see \cite{kripke1971identity} and \cite{maxwell1978rigid}. We understand the capacity of an entity to change the structure of the story in terms of the philosophy of action, see \cite{van1976philosophy}. The structural analysis of narrative first proposed in \cite{shklovsky1925theory} and revived in \cite{propp1968morphology} is mostly focused on a characterization of the action and interaction sequences of 'heroes', the protagonists, and antagonists. \cite{shklovsky1925theory} introduces the concept of 'actor' as an entity that moves the story forward. Thus, AE is understood as the recognition of any 'actor' (no matter hero or not) in the unstructured text of the story. 

It is important to note that we suggest understanding such {\em actionable entity} in the broadest possible terms. For example, if at some point a protagonist opens a window to air the room and the antagonist enters the building through this open window later as the story unfolds, such 'window' could be regarded as a perfect example of an actionable entity. We suggest doing it due to two core reasons. First,  this definition allows us to pinpoint AEs as a broad class of entities mentioned in unstructured texts in natural language. However, the current definition allows further fine-graining of the term depending on the use case or research question. Second, this approach is simple in its naivety and thus allows us to avoid a deeper conceptual discussion of action in the narrative context. We state that if a character interacts with something or someone, we declare it an {\em actionable entity}. The difference between a named entity and an actionable entity is in the interactive nature of the latter. In the next Section, we describe the dataset for AER that we propose. The examples of the dataset clarify these differences further.

\section{Data}

Interactive Fiction games are text-based simulators where a player uses text commands to change the environment and navigate through the story. Such games represent a unique intersection of natural language processing and sequential decision making, making them extremely valuable in a machine learning context and drawing various researchers' attention. For example, \citet{narasimhan2015language} show that one could learn control policies for text-based games with reinforcement learning methods. \citet{cote2018textworld} presents TextWorld, a sandbox learning environment for the training and evaluation of reinforcement learning agents on text-based games. \citet{hausknecht2020interactive} introduce Jericho, a learning environment for human-made interactive fiction games alongside possible benchmarks for language-based autonomous agents. \citet{nelson2006declarative,martin2018dungeons,jain2020algorithmic} and many other authors develop various designs of reinforcement learning agents for interactive fiction that play with various types of feedback and demonstrate various levels of generalization.

Large action spaces impede an agent's learning ability, especially when many actions are redundant or irrelevant. This is especially prevalent in interactive fiction that combines natural language understanding with sequential decision-making. Navigating through interactive fiction, the agent must often detect affordances: the set of behaviors enabled by a situation. Affordance detection benefits domains with large action spaces, allowing the agent to prune its search space by avoiding futile behaviors. \cite{fulda2017can} present such an affordance detection mechanism, and \cite{haroush2018learning} develop the action-elimination architecture that solves quests in the text-based game of Zork, significantly outperforming the baseline agents.

In some sense, Actionable Entities recognition could be linked to affordance detection mentioned above, yet AER is an isolated natural language processing (NLP) task due to its relevance to the structure of the narrative. Further research of AER as a stand-alone NLP task could shed light on fundamental properties of narrative generation, systematic principles of human attention, and structural conditions that these principles impose on story formation. Further in this paper, we illustrate AER applicability to NLP tasks outside the scope of interactive fiction. However, interactive text games are perfect playgrounds to test and benchmark AER algorithms. 

\begin{figure}[h!]\centering
     \includegraphics[scale=0.4]{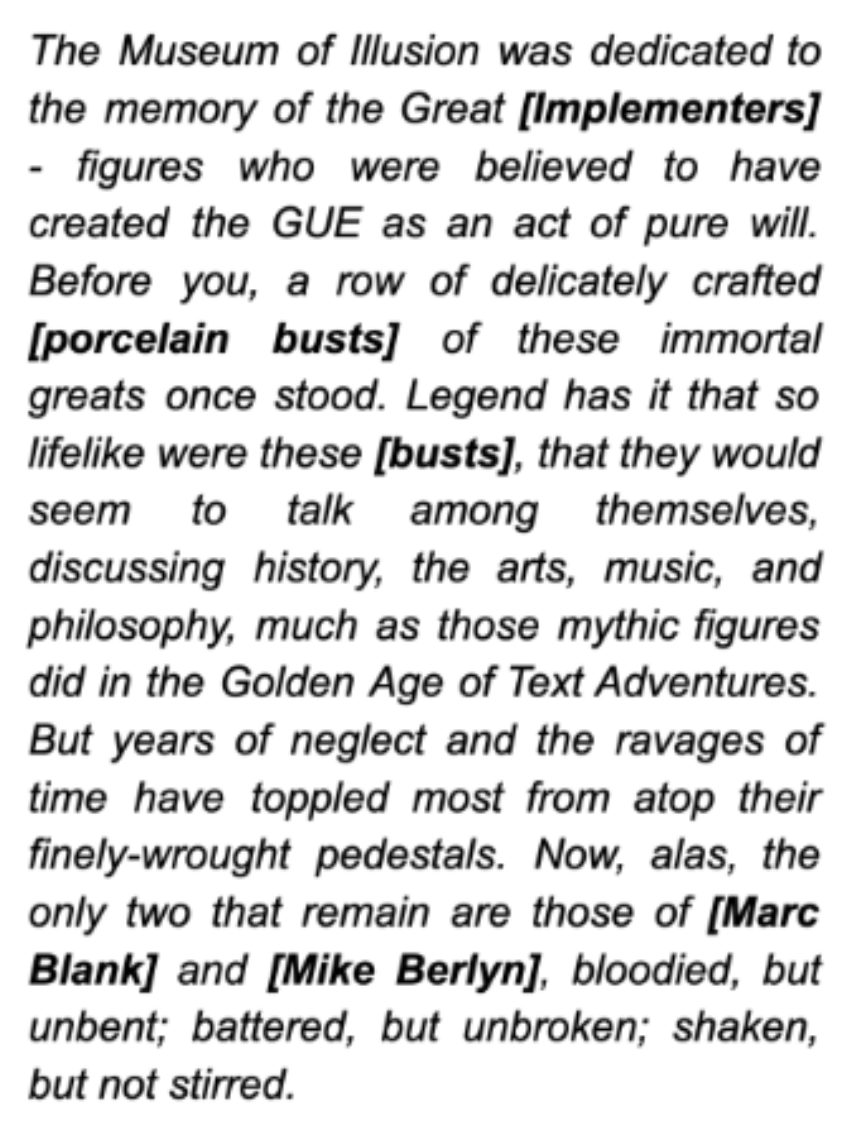}
  \caption{Text description of a location with highlighted items available for interaction. One can clearly see that AER differs from NER though there is an overlap.}
  \label{fig:cor}
\end{figure}

This paper presents a new benchmark — A Benchmark for Actionable Entities Recognition (BAER). The dataset includes 5550 locations with one or more AE in each of them and is based on 995 games. Its' total size is 1.2 megabytes. Every Actionable Entity in the text is labeled as shown in Figure \ref{fig:cor}. The nature of interactive fiction helps to distinguish AER from NER, since there could be several entities in a given location. However, only several of them are actionable and affect the storyline. We publish the resulting BAER dataset\footnote{https://github.com/altsoph/BAER} with labeled AEs and suggest it for AER algorithms benchmarking. 

The data consists of three different types of texts from interactive fiction games and is formatted similarly to the example given in Figure \ref{fig:cor}. The first part of the dataset is based on the Jericho introduced in \cite{hausknecht2020interactive}. It includes 57 interactive fiction games with lists of all locations and all actionable entities. One could assume that the game's authors regarded every actionable entity in these descriptions as something that could help the player move forward in plot evolution. Iteratively searching all such actionable entities in every location with a script, we get a list of Actionable Entities.

We have also collected 24 additional interactive fiction games that have published solutions and are not included in Jericho. For every game, we have created a script that can execute the commands described in the solution and controls the location where the command is executed. As an output of this script, we obtain a list of locations on the critical path that leads to the completion of the game. For every location, we also store a prefix of commands that leads the character to this particular location. Then we brute force the labels of AEs in the following manner. For every location, we reset the game and entered the prefix that brings us into this location. We store the description of the location and start iterating over all objects in this description using the command \verb"examine" $<$\verb"obj"$>$'. If interaction with one of the entities gives a non-trivial reaction from the game, we label this entity as an AE. Most of IF have a standard response to inconsequential actions of the player. If the game's response is different, we assume that the entity is actionable. 

Finally, we include 1500 other interactive fiction games for which we found no working solution. Instead of a solution-based strategy, we implemented a random walk for labeling Actionable Entities in these games. In every game, we did 2500 random steps and implemented the same logic we used for the solution but in a limited number of locations that we were lucky to obtain. 

\section{Task validation}

We propose several ways to approach the AER problem using the BAER dataset and other datasets available in the literature.

\subsection{Contrasting Actionable and Named Entities} \label{contrast}

Let us discuss how AER differs from NER. To do that, we use a T-NER model developed by \citet{ushio2021t}. The T-NER realization based on the XLM-RoBERTa \footnote{https://huggingface.co/asahi417/tner-xlm-roberta-large-all-english} works in the entity span prediction regime marking the beginning and the end of a named entity in the text. The transformer-based architectures significantly outperform classical NLP models on NER tasks, so we see no reason why the situation would differ for AER. Thus, we validate the proposed task regarding the state-of-the-art NER solution. Table \ref{tab:XLM} shows the results of the pre-trained T-NER of the BAER dataset along with results after fine-tuning.

\begin{table}[]\centering
\begin{tabular}{lll}
                  & Accuracy & F1-score \\ \hline
Pre-trained T-NER & 0.05     & 0.00     \\
T-NER fine-tuned on BAER  & 0.50     & 0.51     \\ \hline
\end{tabular}
\caption{Pre-trained XLN-RoBERTa T-NER hardly detects Actionable Entities out of the box. Yet after fine-tuning on BAER for entity span prediction the quality significantly improves.}
\label{tab:XLM}              
\end{table}

Table \ref{tab:XLM}  illustrates that Actionable Entities profoundly differ from Named Entities. Fine-tuning a pre-trained T-NER model on the BAER dataset boosts the F-1 score from virtually zero to around one-half, yet there is room for progress. It is important to note that BAER is far smaller than the NER datasets that are commonly used in the literature. This experiment allows us to conclude that AEs form a distinct category of entities that is out of the scope of current NER: some named entities are actionable, yet many Actionable Entities are NOT named.

Table \ref{tab:relfreq} provides some qualitative understanding of the contrast between named entities and AEs. T-NER classifies detected entities into several categories. After fine-tuning XLM-RoBERTa-based T-NER on BAER dataset, we apply T-NER and our AER model to four different datasets and look for the T-NER categories that AEs dominate. The BAER column stands for the validation part of BAER. There are only three T-NER categories in which Actionable Entities are a majority. These are "product" with only one NER that is not an AE per every eighteen AEs, "person" with one NER that is not an AE per 2.6 AEs, and "work of art" with one NER that is not an AE per two AEs.

\citet{yao2020keep} present a ClubFloyd dataset crawled from the ClubFloyd website\footnote{http://www.allthingsjacq.com/interactive\_fiction.html} and contains 426 human gameplay transcripts, which cover 590 text-based games of diverse genres and styles. The data consists of 223,527 context-action pairs in the format \verb"[CLS]" \verb"observation" \verb"[SEP]" \verb"action" \verb"[SEP]" \verb"next" \verb"observation" \verb"[SEP]" \verb"next" \verb"action" \verb"[SEP]". Every context entry \verb"observation" is accompanied by variants of human attempts to interact with objects and characters in the given context. Table \ref{tab:relfreq} shows that for these context descriptions, T-NER categories "product," "person," and "work of art" are dominated by AEs as well as in BAER. However, there are other AE-dominated categories, such as "chemical." 

\citet{malyshevaDYP} present the TVMAZE\footnote{https://www.tvmaze.com/} dataset. The dataset consists of 13 000 texts that describe plots of TV series split into short episode annotations. WikiPlots\footnote{https://github.com/markriedl/WikiPlots} is a collection of 112,936 story plots extracted from the English Wikipedia. These two datasets are significantly larger than datasets of interactive fiction. Table \ref{tab:relfreq} demonstrates that once again, categories "person" and "product" are dominated by AEs, but such categories as "corporation," "organization," and "group" add up themselves to the picture.

\begin{table*}[ht!]
\centering
\begin{tabular}{ll|ll|ll|ll}
\multicolumn{2}{c|}{BAER}                                                           & \multicolumn{2}{c|}{ClubFloyd}                                                       & \multicolumn{2}{c|}{TV-MAZE}                                                        & \multicolumn{2}{c}{WikiPlots}                                                       \\
\multicolumn{2}{c|}{1 117 texts}                                                    & \multicolumn{2}{c|}{43 795 texts}                                                    & \multicolumn{2}{c|}{299 197 texts}                                                  & \multicolumn{2}{c}{2 070 449 texts}                                                 \\
\hline
\begin{tabular}[c]{@{}l@{}}T-NER\\ category\end{tabular} & $\frac{\nu_{AER}}{\nu_{ner}}$ & \begin{tabular}[c]{@{}l@{}}T-NER\\ category\end{tabular} & $\frac{\nu_{AER}}{\nu_{ner}}$ & \begin{tabular}[c]{@{}l@{}}T-NER\\ category\end{tabular} & $\frac{\nu_{AER}}{\nu_{ner}}$ & \begin{tabular}[c]{@{}l@{}}T-NER\\ category\end{tabular} & $\frac{\nu_{AER}}{\nu_{ner}}$ \\ \hline
product                                                  & 18                       & chemical                                                 & 7.5                      & person                                                   & 36.5                     & product                                                  & 31.2                     \\
person                                                   & 2.6                      & product                                                  & 7.3                      & product                                                  & 12.2                     & corporation                                              & 22.7                     \\
work of art                                              & 2.0                      & person                                                   & 5.5                      & corporation                                              & 3.9                      & person                                                   & 20.1                     \\
organization                                             & 0.7                      & other                                                    & 2.1                      & organization                                             & 3.3                      & group                                                    & 5.4                      \\
location                                                 & 0.5                      & work of art                                              & 1.3                      & chemical                                                 & 3.1                      & chemical                                                 & 4.9                     
\end{tabular}

\caption{Lists of T-NER categories in which Actionable Entities tend to have higher relative frequencies across four different datasets. For every T-NER category, the number of Actionable Entities that fall into this category is divided over  T-NER recognized entities that are not Actionable Entities (provided this number is not zero). The table shows five categories in which Actionable Entities have the highest relative frequency.}
\label{tab:relfreq} 
\end{table*}

An Actionable Entity is a type of entity that plays a causal role in developing a narrative. Since some of such entities are named, modern NER models could detect some percentage of AEs in a text. These would typically be single characters (say, Thor), groups or some organizations (say, Asgard), or objects that are "branded" in some sense (i.e., Mj{\"o}lner, Obedience Potion, or statue of Loki). Along with AEs that are named, a large portion of AEs is not detected within the framework of NER. In this subsection, we have provided a basic benchmark for AER via fine-tuning XLM-RoBERTa based T-NER model and have shown that the model trained on BAER does not have to be only used for interactive fiction but can be applied to other NLP datasets.

\subsection{Actionable Entities and Human World Knowledge}

Humans tend to understand causal relations between entities in a given text intuitively. If a player wants to leave the location, she might try to use the door that was mentioned in the description. If the door is closed, to open it, the player could look for a key, etc. This given understanding of the "world" is crucial for interactive fiction, yet its importance is not limited to textual gameplay. Models endowed with such understanding would be far more usable across the board of NLP tasks. Let us show that AER is a cornerstone for the acquisition of such knowledge. 

In Subsection \ref{contrast}, we have already introduced the dataset presented in \cite{yao2020keep} and used the game context descriptions to explore the differences between NER and AER. Now let us focus on the actions that this dataset contains. These are actual attempts by human players to interact within the proposed environment. One action typically consists of a verb (i.e., go, take, open, etc.) followed by a description of an entity with which the player is trying to interact. We call these entities Action Targets (AT). Since the dataset contains game logs of several players, some ATs could be mentioned several times. These would be the entities that are perceived by humans as more interesting to explore and use in the game. Table \ref{tab:human} summarizes what share of ATs could be labeled by an AER model.

\begin{table*}[ht!]
\centering
\begin{tabular}{l|llll|llll}
             & \multicolumn{4}{c|}{All payer action targets (AT)}                                        & \multicolumn{4}{c}{Unique ATs}                                          \\ \hline
AER model threshold    & p \textgreater 0.5 & p \textgreater 0.65 & p\textgreater{}0.8 & p\textgreater{}0.95 & p \textgreater 0.5 & p \textgreater 0.65 & p\textgreater{}0.8 & p\textgreater{}0.95 \\ \hline
Share of AEs & 0.38               & 0.43                & 0.48               & \textbf{0.57}       & 0.22               & 0.25                & 0.29               & 0.36                \\
that occur in AT list &  &   &     &    &      &   &      &            \\
\hline
Share of ATs & \textbf{0.84}      & 0.65                & 0.30               & 0.05                & \textbf{0.84}      & 0.65                & 0.30               & 0.04                \\ 
labelled as AEs &  &        &          &            &      &   &   &                 \\
\hline
\end{tabular}
\caption{XLN-RoBERTa T-NER fine-tuned on BAER predicts which action targets from the Club Floyd dataset a player would try to interact with. The entities with higher AE probability estimates almost surely end up on the list of action targets. If the entities with lower AE probability are included, they cover up to 84\% of all action targets used by the human players. Since the Club Floyd dataset contains raw interactive fiction data, there are entities with which several players try to interact. The table provides both the results for the full list of action targets as well as for the list of unique action targets.}
\label{tab:human}  
\end{table*}

AER model predicts if an entity is an AE with some probability. Varying this probability threshold would change the prediction of the model. With a higher threshold, the list of AEs would naturally be shorter. Table \ref{tab:human} shows that the share of AEs that overlap with some human action targets grows if the threshold is higher. At the same time, if the model has a lower threshold, it predicts more than 80\% of all the entities that humans try to interact with.

\section{Actionable Entities and the structure of narrative}

\citet{papalampidi2019movie} introduce a  TRIPOD dataset\footnote{https://github.com/ppapalampidi/TRIPOD} that includes plot synopses with turning point annotation. They also suggest looking at five turning points in the script since this variant is commonly employed by screenwriters as a practical guide for producing successful screenplays. These five turning points are:
\begin{itemize}
    \item Opportunity — the introductory event that occurs after
the presentation of the setting and the background of the main characters;
    \item Change of Plans — the event where the main goal of the story
is defined;
    \item Point of No Return — the event that pushes the main character(s) to commit to their goal fully;
    \item Major Setback — the event where everything falls apart;
    \item Climax — the final event of the main story.
\end{itemize}

\begin{figure}[h!]
    \includegraphics[scale=0.38]{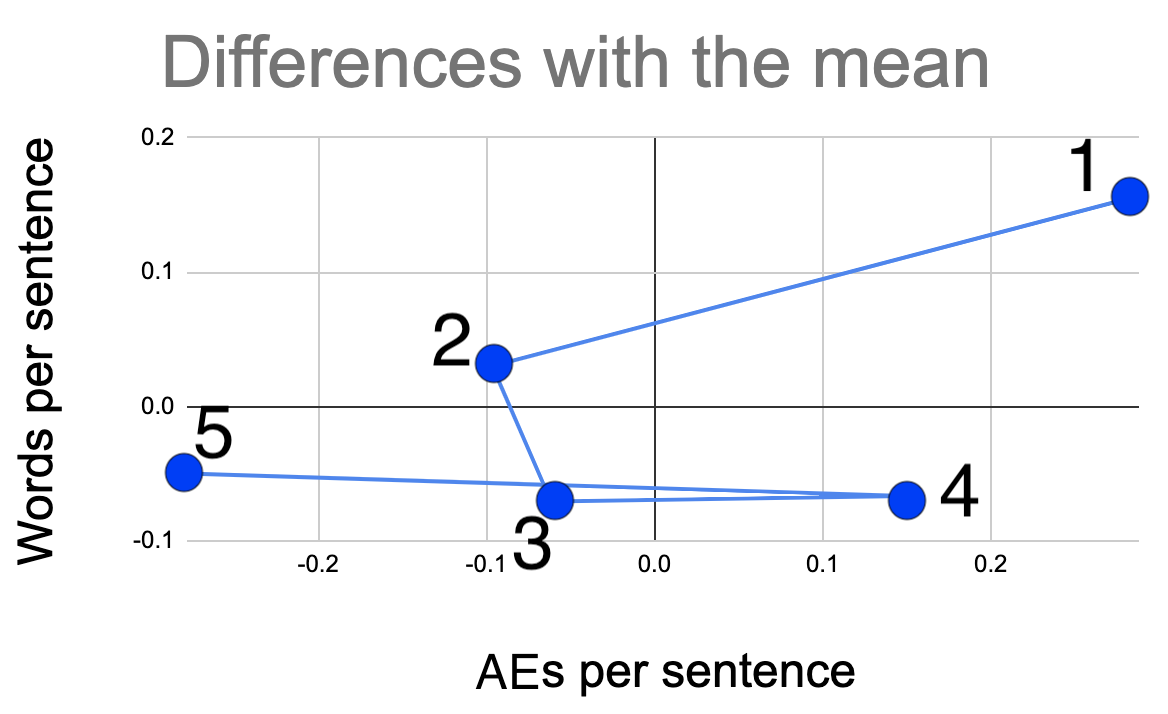}
  \caption{Average number of AEs per sentence and the average number of words in a sentence differ in different turning points of the story. The axis shows the difference between these values calculated at the corresponding turning point and the story at large. The first turning point of a story has more words and AEs per sentence than usual. Then the sentences tend to get shorter. The story's second and third turning points have fewer Actionable Entities per sentence, while at the fourth turning point, the number of AEs per sentence peaks.}
  \label{fig:tripod}
\end{figure}

Figure \ref{fig:tripod} shows how the average number of AEs per sentence varies in these turning points. Since we are interested in relative dynamics within a story, we calculate the average number of AEs per sentence and see its difference with an average number of AEs across all five turning points. Since the number of AEs per sentence might be higher for longer sentences, we also calculate how the average number of words per sentence differs at every turning point.

Turning point number one — Opportunity — tends to have more words per sentence and more Actionable Entities in them. As the plot thickens in Change of Plans and Point of No Return, sentences get shorter. However, the number of Actionable Entities per sentence gets lower as well. In his letter to Lazarev, Anton Chekhov famously wrote: "One must never place a loaded rifle on the stage if it is not going to go off. It is wrong to make promises you do not mean to keep." Rephrasing Chekhov's quote, one could say that some of the "loaded rifles" are already placed on the stage, and we are waiting for the shots. Indeed, in Major Setback number of AEs per sentence jumps, while the average number of words per sentence remains low. In Climax, the number of AEs drops to its lowest.

\begin{table}[]\centering
\begin{tabular}{r|rrrrr}
\multicolumn{1}{l|}{\begin{tabular}[c]{@{}l@{}}Turning \\ Point \#\end{tabular}} & 1      & 2      & 3      & 4      & 5      \\ \hline
1                                                                                & 32.2 & 7.2  & 6.3  & 6.8  & 5.4  \\
2                                                                                & 0.0  & 21.7 & 4.0  & 4.0  & 3.3  \\
3                                                                                & 0.0  & 0.0  & 17.3 & 2.5  & 2.6  \\
4                                                                                & 0.0  & 0.0  & 0.0  & 16.0 & 2.8  \\
5                                                                                & 0.0  & 0.0  & 0.0  & 0.0  & 12.8
\end{tabular}
\caption{Percentage of AEs in every turning point of a story. Every row represents AEs that first occurred in the corresponding turning point and then reoccurred in later parts of the story that stand in the corresponding column. The diagonal sums to 100\% representing all first occurrences.}
\label{tab:fo}  
\end{table}

Table \ref{tab:fo} shows how AEs first occur and reoccur in the story depending on the turning point. Indeed, almost one-third of AEs first emerge in the Opportunity part of the story. In every later turning point, five to seven percent of all AEs in a story are the ones that we first meet in the Opportunity part. The percentage of first occurrences consistently drops from 32.2\% in the first to 12.8\% in the last turning point. 

This is a qualitative picture that illustrates the potential of AER as a stand-alone NLP task: since Actionable Entities affect causal relations within the narrative, they could be useful for the overall analysis of the narrative structure.

\section{Discussion}

In recent years we have seen various exciting results in the area of Natural Language Generation (NLG). One line of research addresses the generation of semi-structured texts varying from dialogue responses, see \cite{li2016deep,li2017adversarial,li2019dialogue}, to traditional Chinese poetry, see \cite{zhang2014chinese,wang2016chinese}. Another line of research addresses the generation of stylized texts, see \cite{tikhonov2018guess,yang2018stylistic}. There have been recent results that try to generate longer blocks of text, such as \cite{kedziorski2019understanding} or \cite{agafonova2020paranoid}, yet the generation of longer texts is only possible under certain stylistic and topical constraints that exclude the problem of narrative generation altogether.

We believe understanding narrative and its key principles is paramount to further advances in NLG. However, the narrative is fundamentally causal. Thus, a deeper understanding of causality in NLP should provide new insights into the structure of the narrative. Currently, our understanding is lacking due to a variety of reasons:
\begin{itemize}
    \item human cognition is often verbal and narrative-based, which makes attempts to conceptualize narrative implode on themselves;
    \item narration is not only a linguistic but a cultural act that fundamentally affects humans as 'cultural animals'. This influence of narrative hinders its conceptualization in rigorous mathematical terms;
    \item narrative is centered around long-term dependencies that could be formally characterized as a critical behavior of language, see \cite{lin2017critical}.
\end{itemize}

These very general issues need particular attention, but we believe the AER is an exemplary step to address these problems. First, AER allows addressing the notion of 'actor' in line with \cite{shklovsky1925theory}, broadening it from a person to any entity that interacts with a storyline. This ability to personalize various entities could be regarded as a cornerstone associative mechanism that might be a basis for creative cognition \cite{mednick1962associative}. Secondly, as with any NLP task, AER could be developed for every culture or language. Further development of AER could provide general cross-cultural insights into the process of story formation. Finally, AEs significantly contribute to the criticality of language in line with ideas expressed in \cite{lin2017critical}. AEs reoccur in the story, and understanding which object is pivotal for the narrative to unfold is one of the key bottlenecks to automated narrative generation. 

In an interactive setting, trying to interact with the world can help to learn AEs. At the same time, the notion of AE can help models discover affordances in games. One could also look into how large language models perform in a zero-shot manner on AER tasks. We believe that all these factors make AER an NLP task that could bring us further toward narrative generation and provide insights into interactive learning. 

\section{Conclusion}

This paper presents a new causal natural language processing task — Actionable Entities recognition (AER). It formalizes the notion of an Actionable Entity as any entity that could be engaged by one of the protagonists and thus affect the story's development in line with the structuralist sense. It provides a corpus of labeled texts to introduce the benchmark for AER. To illustrate the differences between AER and NER, it validates the proposed task on four different datasets. The paper demonstrates that the AER model predicts human attempts to interact with entities within interactive fiction. Finally, it shows how the AER classifier could be applied to an external dataset to obtain insights into the structure of the narrative.

\section{Acknowledgments}
Ivan Yamshchikov obtained some of the results while working at LEYA Laboratory by Yandex and Higher School of Economics in St. Petersburg. His work is an output of a research project implemented as part of the Basic Research Program at the National Research University Higher School of Economics (HSE University).

\bibliographystyle{acl_natbib}
\bibliography{gun}

\end{document}